%% file: main.tex
\documentclass[10pt,twocolumn,letterpaper]{article}

\usepackage[pagenumbers]{wacv}



\input{preamble}
\definecolor{wacvblue}{rgb}{0.21,0.49,0.74}
\usepackage[pagebackref,breaklinks,colorlinks,allcolors=wacvblue]{hyperref}

\def\wacvPaperID{609} %
\def\confName{WACV}
\def\confYear{2026}

\title{CADE: Continual Weakly-supervised Video Anomaly Detection with Ensembles}

\author{
Satoshi Hashimoto \quad
Tatsuya Konishi \quad
Tomoya Kaichi \quad
Kazunori Matsumoto \quad
Mori Kurokawa\\
KDDI Research, Inc.\\
}

\usepackage{multirow}
\usepackage{mathtools}
\usepackage{adjustbox} %

\begin{document}
\maketitle
\input{sec/0_abstract}    
\input{sec/1_intro}

\input{sec/2_related}
\input{sec/3_proposed}
\input{sec/4_exp}
\input{sec/5_conc}
{
    \small
    \bibliographystyle{ieeenat_fullname}
    \bibliography{main}
}

\clearpage
\maketitlesupplementary
\input{sec/6_supp}

\end{document}

%% file: sec/0_abstract.tex
\begin{abstract}
Video anomaly detection (VAD) has long been studied as a crucial problem in public security and crime prevention. In recent years, weakly-supervised VAD (WVAD) has attracted considerable attention due to their easy annotation process and promising research results. 
While existing WVAD methods mainly tackle static datasets, the possibility that the domain of data can vary has been neglected. To adapt such domain shift, the continual learning (CL) perspective is required because otherwise additional training using only newly incoming data could easily cause performance degradation for previous data, i.e., forgetting. Therefore, we propose a brand-new approach, called \underline{C}ontinual \underline{A}nomaly \underline{D}etection with \underline{E}nsembles (CADE) that is the first work combining CL and WVAD viewpoints.
Specifically, CADE uses the Dual-Generator (DG)
to address data imbalance and label uncertainty in WVAD. 
We also found that forgetting exacerbates the ``incompleteness'' where the model becomes biased towards certain anomaly modes, leading to missed detections of various anomalies.
To address this, we propose a Multi-Discriminator (MD) ensemble that captures missed anomalies in past scenes due to forgetting, using multiple models.
 Extensive experiments show that CADE significantly outperforms existing VAD methods on the common multi-scene VAD datasets, such as the ShanghaiTech, UCF-Crime, and Charlotte Anomaly Dataset.
\end{abstract}

%% file: sec/1_intro.tex
\section{Introduction}
\label{sec:intro}
To ensure the safety and orderliness of society, Video Anomaly Detection (VAD) has been a subject of extensive research in the field of computer vision. Those studies aim to identify various anomalies, such as assault, traffic accidents, and theft, within surveillance video footage. With the increasing prevalence of surveillance cameras from a public security perspective, the manual monitoring of video feeds has become inefficient and highly challenging, raising the demands of automatic camera monitoring technology.
VAD approaches can be grouped into two categories: unsupervised approaches using only normal data ~\cite{liu2018future,yu2022deep,nguyen2019anomaly,liu2021hybrid,ravanbakhsh2017abnormal,ionescu2019object,zaheer2022generative,doshi2022rethinking} and weakly supervised VAD approaches (WVAD) using weak annotations at the video level ~\cite{feng2021mist,zhang2023exploiting,sultani2018real,zhong2019graph,tian2021weakly,zhang2019temporal,lv2023unbiased,wan2020weakly,wu2020not}. 
Especially, WVAD only requires lower annotation cost, thus it gains a lot of attention in recent years.
Existing WVAD, however, do not assume the case where the domain of data can vary upon several factors rapidly changing in environment such as time, climate, or location, and models need to adapt the new domain's data continuously. While additional training is essential to deal with such domain shift, saving all the data (i.e., not only the new domain's data but also previous domain's data) may be prohibited due to privacy concerns.

\begin{figure}[t]
\centering
\begin{minipage}[b]{\linewidth}
  \centering
  \includegraphics[width=\textwidth]{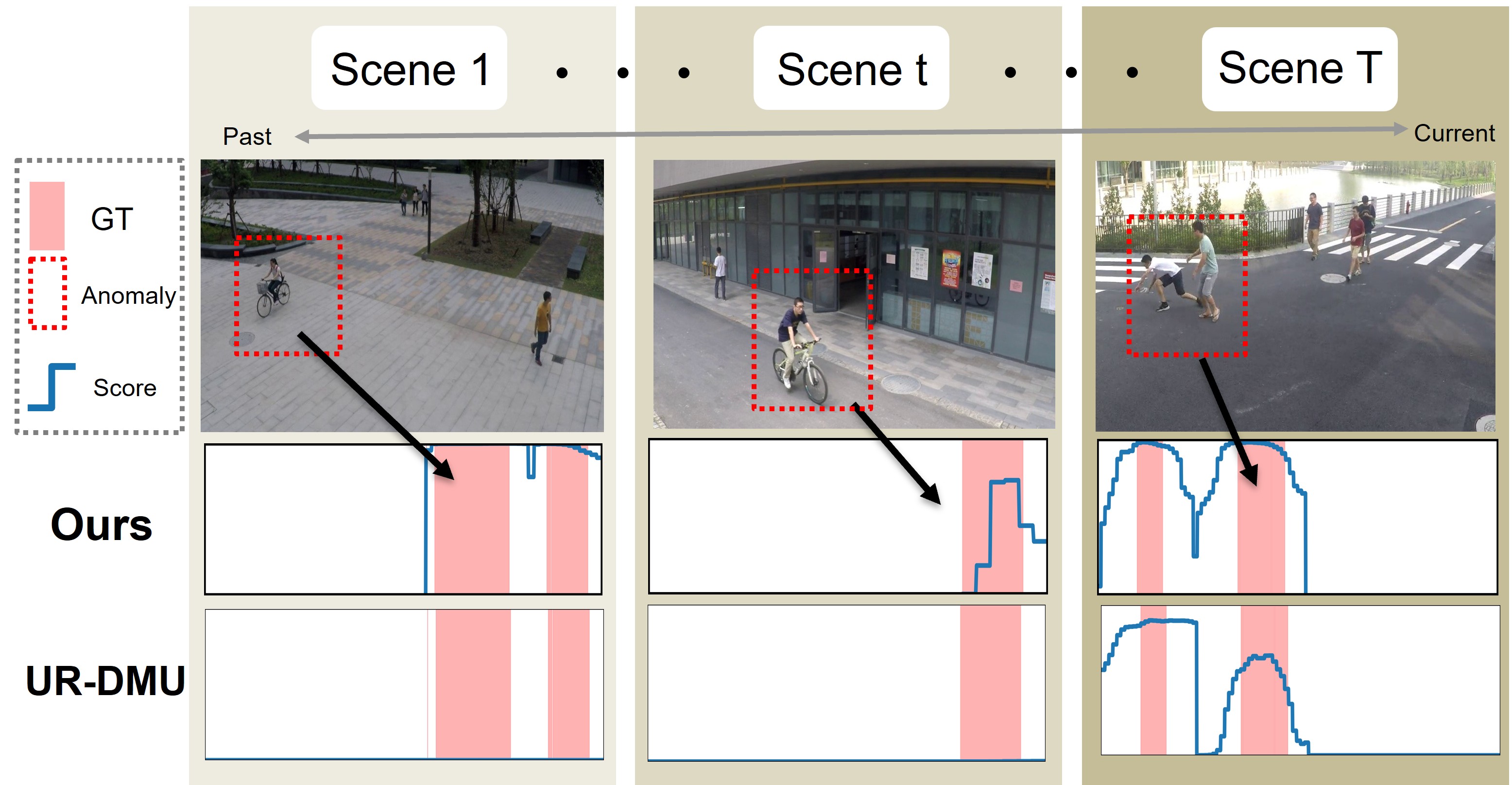}
\end{minipage}
\caption{%
Our proposed method CADE is robust to additional training and maintains detection accuracy. While existing UR-DMU causes forgetting as its score becomes inactive when learning multiple scenes, CADE successfully indicates anomalies. }
\label{fig1}
\end{figure}

Nevertheless, a simple additional training, i.e., fine-tuning, only with new domain's data leads to the performance degradation on previous domains, which is known as catastrophic forgetting in the continual learning (CL) research. %
While CL involves several setups for different situations,
anomaly detection can be described as a domain incremental learning (DIL) problem 
that tackles covariate-shifted data where input data coming from different domains have the same set of labels. CL can also assume other constraints; for instance, after the learning of a domain finishes, the domain's data is inaccessible when learning a new domain.
Among existing CL approaches especially for DIL, generative replay (GR) is one of the most promising techniques~\cite{hayes2021replay}. %
GR consists of two components: generator and classifier.
Generator generates past samples and provides them to the classifier, which allows the classifier to learn not only the current domain but also the past ones, which suppresses forgetting the previous knowledge. %
This GR strategy is advantageous in terms of memory efficiency and privacy protection as it does not require storing real past samples.
Despite the success of GR, we observed that naively applying existing GR to WVAD does not reach sufficient performance %
because of the following two problems. Firstly, GR is vulnerable for imbalanced data and labeling noise. In the case of anomaly detection, normal data is expected to be dominant in terms of data quantity, and in WVAD, GR is also affected by labeling noise due to weak labels. These can result in generated samples being biased towards normality, making it difficult to generate appropriate anomaly samples. %
Secondly, conventional GR approaches utilize a single generator only to prevent the classifier from forgetting. 
However, they still suffer from forgetting because the structure of a single generator and discriminator can promote bias towards certain anomaly modes.
Furthermore, we found that the issue of incompleteness, characterized by biased outputs --- an inherent challenge in WVAD --- becomes even more prominent due to catastrophic forgetting in CL. An example of this is shown in \cref{fig1}. 
When existing WVAD methods are directly applied to CL scenarios, outputs become biased towards the current scene, leading to an increase in missed detections in past scenes.

To solve those problems, we propose a brand-new approach, called CADE (Continual Anomaly Detection with Ensembles), that solves WVAD from the CL perspective by introducing the novel module connections working with multiple discriminators. To our knowledge, CL-aware WVAD has not been done before.
Unlike existing methods, our proposed CADE exploits GR to mitigate forgetting inside VAD's anomaly discriminators.
We also prepare Dual-Generator (DG), a normal and an anomaly generator pair, that enhances the model's robustness to the imbalance between normal and abnormal based on the video-level annotations. Additionally, by ensembling the outputs of Multi-Discriminator (MD), we address the incompleteness of anomaly detection.
We show that fusing the outputs of multiple models enhances prediction quality.
In summary, the contributions of this paper are as follows:
\begin{itemize}

\item This is the first paper that utilizes CL techniques to allow WVAD models to perform with less forgetting for new coming data from different domains. We do not just apply generative replay (GR) to WVAD but also propose a novel architecture that addresses the drawbacks (i.e., vulnerable to imbalanced data) of GR appearing in the scenarios of WVAD.
\item Our proposed method CADE is able to detect anomalies efficiently and effectively by ensembling multiple discriminators used for replay during inference. This strategy
contributes to overcoming the imperfections of anomaly classifiers in WVAD.
\item Our proposed method is effective for any WVAD methods that use classification models. 
\item Extensive experiments show that CADE significantly outperforms existing VAD methods on the common multi-scene VAD datasets, such as the ShanghaiTech, UCF-Crime, and Charlotte Anomaly Dataset.
\end{itemize}

%% file: sec/2_related.tex
\section{Related Work}
\label{sec:related}

\subsubsection{Video Anomaly Detection}
Approaches to video anomaly detection can be divided into three main categories: 1) Unsupervised Video Anomaly Detection (UVAD) and 2) Weakly-supervised Video Anomaly Detection (WVAD). 3) Large Language Models (LLM) based methods.  UVAD is motivated by the needs to address the difficulty of collecting supervised data for anomaly detection. 
Instead of actual collected data,
based on the assumption that the training data covers all modes of normal, 
dictionary learning using only normal data~\cite{ren2015unsupervised}, reconstruction-based methods~\cite{liu2018future,yu2022deep,liu2021hybrid,ravanbakhsh2017abnormal,ionescu2019object,zaheer2022generative,nguyen2019anomaly}, and one-class classifier-based methods have been proposed. 
In particular, reconstruction-based methods using deep auto-encoders have been successful%
;Ravanbakhsh et al.~\cite{ravanbakhsh2017abnormal} used pix2pix~\cite{isola2017image} in their model to learn the task of domain transformation between optical flow and raw images, and proposed an anomaly detection method using the difference between the input and output. 
A major drawback of UVAD, however, is that it depends on a strong assumption that all normals are covered in the training data, while it cannot be guaranteed. Once unprecedented normal data come as input, VAD systems suffer from false positives. 
Additionally, the UVAD framework that lacks an explicit definition of anomalies may also lead to the occurrence of false positives.
Furthermore, although several methods of Fully-Unsupervised Video Anomaly Detection ~\cite{yu2022deep,zaheer2022generative} have been proposed in recent years, they cannot be used in surveillance applications that require the detection of complex anomalies because they depend on generative models that are supposed to handle only simpler anomalies.

On the other hand, WVAD methods adopt a discriminative technique to capture complex anomalies, 
as described in the introduction section, becoming a hot topic in the VAD field recently.
WVAD was first proposed in the Multiple Instance Learning (MIL) ranking framework using weak annotation at video level, not frame level~\cite{sultani2018real}.
In MIL of WVAD, a video and video segment are defined as a bag and an instance, respectively.
WVAD aims at the frame-level anomaly detection only from video-level annotation indicating whether at least an anomaly exists in a given video, which significantly reduces annotation cost.
In particular, based on weak annotations, normal and anomaly bags (i.e., videos) are treated as negative and positive bags, respectively.
Next, a ranking strategy (e.g., top-k) relies on scores in each bag to select an instance from the positive bag that is most likely anomaly and another instance from the negative bag that is most near to anomaly.
Inspired by Sultani et al.'s work~\cite{sultani2018real}, 
a number of MIL-based methods have been proposed~\cite{feng2021mist,zhang2023exploiting,sultani2018real,zhong2019graph,tian2021weakly,zhang2019temporal,wu2020not,wan2020weakly}. Although they achieved success in their problem setting, it has been unveiled that WVAD has other issues, called uncertainty and incompleteness. 
Zhong et al.~\cite{zhong2019graph} took WVAD as a learning task with noisy labels and proposed a noise refinement module using Graph Convolutional Networks (GCN). The noise arises from 'uncertainty' caused by incorrect labeling, where normal instances may be mislabeled as anomalies within positive bags when video-level labels are directly applied to instances.
Feng et al.~\cite{feng2021mist} propose a two-stage learning strategy that first generates pseudo labels for an instance-level classifier and then trains only the classifier. %
This strategy allows to directly obtain video discriminative representations using instance-level pseudo-labels.
Zhang et al.~\cite{zhang2023exploiting} point out that although existing MIL setting assumes that a positive bag has only one anomaly instance and thus depends only on it for judge, the bag may contain multiple instances, which leads to biased anomaly detection. 
Consequently, the classifier is prone to triggering missed detections, leading to incompleteness.
As aforementioned, although WVAD has achieved some success, its application to surveillance applications requires additional sequential training of the model to handle dynamic data; since existing WVAD approaches only assume static datasets for evaluation, thus this remains an essential issue.
However, because WVAD approaches assume a single, static dataset, they face major challenges such as catastrophic forgetting, leading to incomplete anomaly detection when continuously learning additional data in a potentially variable environment.
In contrast, CADE overcomes the incompleteness of anomaly detection through the ensemble of multiple discriminators.

Meanwhile, inspired by the recent success of Multi-modal LLM, several applications have emerged in the VAD field~\cite{Chen_2024_CVPR,Du_2024_CVPR,Zhang_2024_HolmesVAD,lv2024videoanomalydetectionexplanation}. Considering practical use in real-world surveillance applications, it is natural to consider fine-tuning (FT) for specific tasks, and indeed, FT-based approaches have shown significant results. 
However, these methods also do not account for real-world data fluctuations, such as variations in location, scene, weather and season, are subject to forgetting due to task-specific adaptation.

The Most similar to our work is the study done by Doshi et al.~\cite{doshi2022rethinking}, who propose a UVAD scenario-based continual learning method.
However, the method lacks comprehensive validation and UVAD approaches suffer from false-positives, limiting their applicability to surveillance.
Zhu et al.~\cite{zhu2022towards} extend WVAD to Open Set tasks with both known and novel anomalies. However, from the perspective of CL, their framework does not deter forgetting about past data. 
\subsubsection{Generative Replay}
The purpose of CL is to suppress catastrophic forgetting of past tasks (in our VAD case, tasks correspond to domains) when a new task is learned. CL methods do not use full of past data to train the model again as they assume the situation where it is difficult to hold the entire past data forever due to the storage constraints or privacy concerns.
Replay-based approaches~\cite{rebuffi2017icarl} that exploit memory buffer to store a small amount of past data and use it during learning a new task have gained much attention from practitioners due to its promising performance and simple implementation. A line of research on replay in CL has shown that the previous data is useful to mitigate forgetting; however, these approaches are not privacy compliant because they store real data in memory.

Besides actually storing samples, methods using generative replay (GR)~\cite{pmlr-v202-gao23e,ye2020learning,shin2017continual,hayes2021replay} are proposed that prepare a generator inside a model for past tasks and generate samples as substitute of real past samples.
In particular, GR consists of two parts: a classifier and a generator. The generator, indeed a GAN model, produces past samples and feeds them to the classifier, which allows the classifier to learn not only the current task but also past tasks even if the model has access only to the current task data, contributing to preventing forgetting. 
Shin et al.~\cite{shin2017continual} proposed a GR method using WGAN-GP as generator. And 
Gao et al. \cite{pmlr-v202-gao23e} 
proposed a new diffusion-based model that introduces a new connection flowing from classifier to generator.
Wiewel et al. \cite{wiewel2019continual} proposed a method that is closest to our motivation, applying VAE-based generative replay to the anomaly detection task.
However, all these methods assume an ideal setup where classes are clearly separable. In more realistic WVAD scenarios, anomaly labels are uncertain due to weak annotations, and the amount of anomaly data is limited, leading to imbalanced datasets. Consequently, simply applying GR methods may cause the generator to bias towards normal data, which could hinder the effective functioning of replay. Additionally, the structure of a single generator-classifier pair may not allow the powerful capabilities of the classifier to fully contribute to the GR system, especially in complex scenarios like WVAD.
In contrast, CADE addresses imbalanced data with dual generators corresponding to normal and anomaly classes, along with multiple discriminators.

\subsubsection{Ensemble}
It has been shown that ensembles improve performance by fusing the prediction results of multiple neural networks, thereby adding diversity to the solution~\cite{doan2023continual}. %
In anomaly detection, generative adversarial networks (GAN) ensemble method~\cite{han2021gan} was proposed that prepares
multiple generators and discriminators in a GAN framework ~\cite{goodfellow2014generative} 
and lets them be jointly trained and used together in inference, showing significant performance improvement.
In the context of CL, there is a lack of research focusing on the use of ensemble models, and only a few methods have been proposed by Doan et al.~\cite{doan2023continual}. They show that ensemble is also effective in the context of CL, but point out the bottleneck of multiple deep neural networks, which increases the cost of preparing multiple models.
CADE incorporates the GAN discriminator used in GR into the ensemble, avoiding the need to create multiple independent models.

In summary, most existing VAD methods cannot handle additional learning because they do not take into account the dynamic environment. GR methods for CL are inefficient because they are not suited to the uncertain and imbalanced data in WVAD. Furthermore, existing ensemble methods face the cost-related challenge of preparing multiple models individually. The proposed method solves these problems simultaneously.

%% file: sec/3_proposed.tex
\section{Proposed Method}
\label{sec:proposed}

\begin{figure*}[t]
\centering
\includegraphics[width=\textwidth]{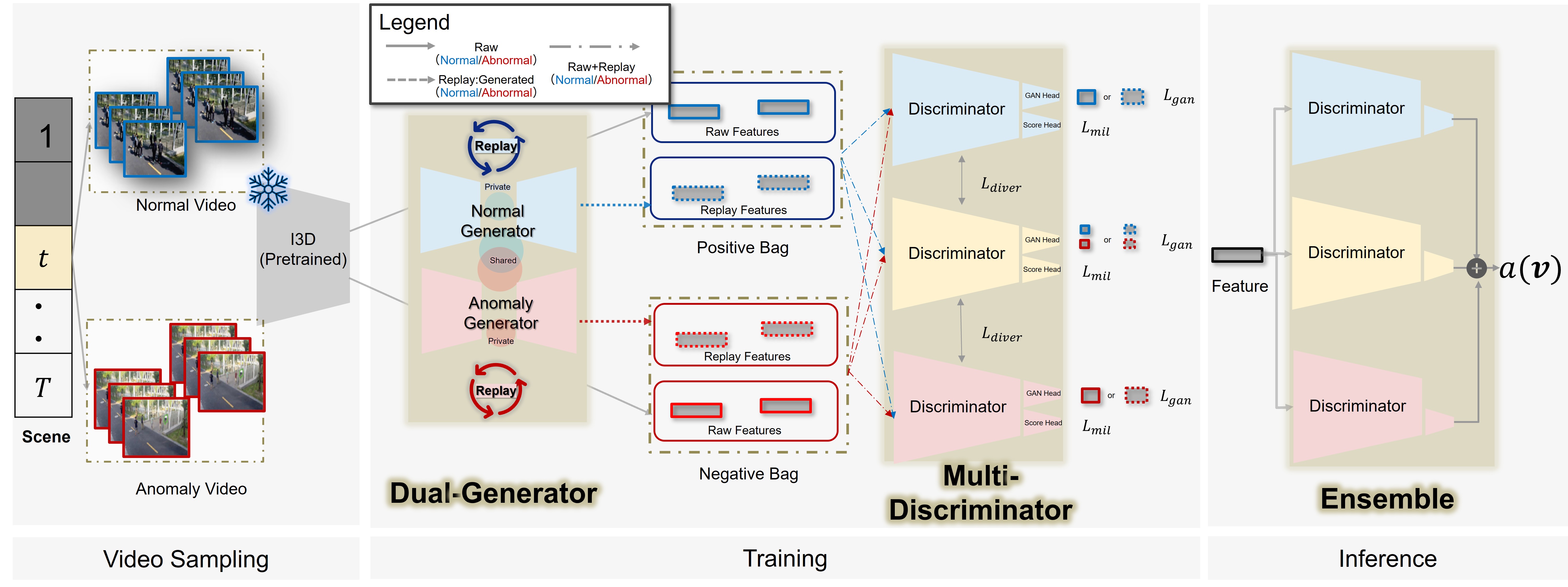}
\caption{Our CADE consists of three key components: Dual-Generator, Multi-Discriminator, and an inference-time ensemble. Following the DIL setup, the model continuously learns using data divided by scene (domain). CADE can be easily integrated with existing WVAD methods by incorporating Dual-Generator and paired discriminators.}
\label{fig2}
\end{figure*}

We provide the detailed explanation of our CADE architecture. As illustrated in the overview in Figure 2, CADE consists of three key components: \textbf{Dual-Generator (DG)} that addresses the imbalance and uncertainty in WVAD data, \textbf{Multi-Discriminator (MD)} that overcomes the incompleteness of anomaly detection, and the \textbf{inference-time ensemble} that efficiently mitigates forgetting. 
Although CADE can be applied not only to MIL but to any WVAD methods, in this paper, we formulate CADE in the MIL setup as MIL is the most common approach in WVAD.

\noindent\textit{Notation.} Let a video $V$ be split into non-overlapping segments (instances). A bag $B(V)=\{\mathbf{v}^1,\dots,\mathbf{v}^{|B|}\}$ is associated with a video-level weak label $y\in\{0,1\}$ ($y{=}1$ anomalous). Instance features are extracted by a frozen backbone $\mathcal{I}$ (e.g., I3D/C3D) as $\mathbf{f}^i=\mathcal{I}(\mathbf{v}^i)\in\mathbb{R}^K$. The discriminator’s anomaly score head outputs $s(\mathbf{f})\in\mathbb{R}$ for each instance. All adversarial objectives are defined in the feature space $\mathbf{f}$. %

First, following the continual learning setup, the training dataset is divided into $T$ domains, and the MIL model discriminator $\mathcal{D}$ is optimized for each successive arrival of a new domain's data. 
In our CADE, we use DMVAEGAN, an extension of DMVAE~\cite{Lee_2021_CVPR} to GAN, as DG for normal and anomaly data to optimize the dataset in domain $t \enspace (1\le t \le T)$, while at the same time letting it produce replay data to the discriminator to prevent forgetting.

The discriminator has a GAN~\cite{goodfellow2014generative} header in addition to the header that outputs the anomaly score. 
Thus, MD ($\mathcal{D}$, $\mathcal{D_\text{n}}$, and $\mathcal{D_\text{a}}$) and DG ($\mathcal{G_\text{n}}$ and $\mathcal{G_\text{a}}$) can mutually optimize each other through adversarial learning. 
In the following, we first formulate the task of WVAD in DIL, and then describe our method by component and procedure such as generator, discriminator, and ensembling.

\subsection{Formulation of MIL in DIL}
In this section, we formulate MIL problem in the form of DIL, where data is split into several domains and has the same set of labels.
Each video belongs to a domain $t$ and annotated with a video-level binary label (i.e., weak label) indicating whether the video includes anomaly events or not.
Note that in WVAD, frames in a video are not annotated; instead, only video-level labels are available during training. In the testing phase, the goal of WVAD is to estimate a frame-level anomaly score that indicates which frame involves an anomaly event.

Following prior MIL-based methods~\cite{feng2021mist,zhang2023exploiting,sultani2018real,zhong2019graph,tian2021weakly,zhang2019temporal}, we split a video into non-overlapping video segments, each of which is defined as instance. 

Based on the weak labels (i.e., video-level labels), we denote a negative (normal) bag by $B^{-}$ when $y{=}0$ and a positive (anomalous) bag by $B^{+}$ when $y{=}1$, with sizes $|B^{-}|$ and $|B^{+}|$, respectively. %
Each video segment is transformed into a $K$-dimensional feature vector, $ \mathbf{f} = \mathcal{I}(\mathbf{v})$, through a pretrained 3D convolutional network (e.g., I3D~\cite{carreira2017quo}/C3D~\cite{tran2015learning}), $\mathcal{I}$.

In our method, finally, the features of a negative bag 
and a positive bag are denoted by $\tilde{B}^{-} = \{\mathbf{f}^{i}\}_{i=1}^{|B^{-}|}\cup\{\tilde{\mathbf{f}}^{i}\}_{i=1}^{r^{-}}$ and $\tilde{B}^{+} = \{\mathbf{f}^{i}\}_{i=1}^{|B^{+}|}\cup\{\tilde{\mathbf{f}}^{i}\}_{i=1}^{r^{+}}$, respectively, where the pseudo-features $\tilde{\mathbf{f}}$ are replayed by the dual-generators ($\mathcal{G}_{\text{n}}$, $\mathcal{G}_{\text{a}}$) with fixed replay ratios $r^{-},r^{+}$. That is, the discriminator is trained with a mixture of real feature instances and pseudo-feature instances. %

Following prior work~\cite{sultani2018real}, we assume that the segment with the highest anomaly score in the positive bag is most likely to be a true positive instance and introduce the following Deep MIL ranking objective function: 
\begin{equation}
\max_{\mathbf{u} \in \tilde{B}^{-}} s(\mathbf{u}) \;<\; \max_{\mathbf{v} \in \tilde{B}^{+}} s(\mathbf{v}) .
\end{equation}
More specifically, the following MIL ranking loss is employed in order to maximize the separability between positive and negative instances~\cite{sultani2018real}.
\begin{equation}
\mathcal{L}_\text{MIL} \;=\; \max\!\left(0,\, 1 - \max_{\mathbf{v} \in \tilde{B}^{+}} s(\mathbf{v}) + \max_{\mathbf{u} \in \tilde{B}^{-}} s(\mathbf{u}) \right).
\end{equation}

Also, for our interest to suppress forgetting, we use a replay generator.
The replay generator $\mathcal{G}$ and discriminator $\mathcal{D}$ are trained in the GAN framework. In the GAN, the discriminator learns to distinguish between samples deriving from two data distributions, while the generator learns to mimic the actual distribution as much as possible. Being inspired by GAN, we judge the quality of sample generation of the generator based on the discriminator and feedback it to the generator. 

In the GAN framework, the objectives of $\mathcal{G}$ and $\mathcal{D}$ networks are described below (feature space):
\begin{equation}
\begin{split}
\min_{\mathcal{G}} \max_{\mathcal{D}} V(\mathcal{D}, \mathcal{G}) &= \mathbb{E}_{\mathbf{f}\sim p_{\text{data}}(\mathbf{f})}[\log \mathcal{D}(\mathbf{f})] \\
&\quad + \mathbb{E}_{\mathbf{z}\sim p_{\mathbf{z}}}[\log(1 - \mathcal{D}(\mathcal{G}(\mathbf{z})))]. 
\end{split}
\end{equation}
where $\mathbf{z}$ is random noise and $\mathbf{f}$ denotes real features; $\mathcal{G}$ tries to minimize the objective while $\mathcal{D}$ tries to maximize the objective. 

\subsection{Dual-Generator}
As discussed in the introduction, existing GR methods are not well-suited for the uncertain and imbalanced data in WVAD. Therefore, we extend and introduce DMVAE~\cite{Lee_2021_CVPR}, considering the characteristics of WVAD data, to separate normal and abnormal classes and achieve superior latent representations in the observation space. This method consists of DG, $\mathcal{G_\text{n}}$ and $\mathcal{G_\text{a}}$, to handle normal and abnormal classes. In DMVAE, each class has a unique private latent space and a shared latent space, effectively capturing the representations of rare anomalies and overlapping representations across classes. Such multi-faceted trait of DMVAE is expected to capture the complex distributions over normal and abnormal classes more effectively than existing GR methods~\cite{gulrajani2017improved,shin2017continual} that use only a single model such as WGAN-GP.
Furthermore, similar to existing GR methods~\cite{gulrajani2017improved}, we adopt the GAN architecture and add discriminators,$\mathcal{D_\text{n}}$,$\mathcal{D_\text{a}}$ , to monitor the quality of each generator.
The input/output space of the model is the feature space extracted from a pretrained model such as I3D.

\noindent
\textbf{Optimization}:
The objective of each generator is as follows:
\begin{multline}
\mathcal{L} \;=\;
\mathbb{E}_{q(\mathbf{z}\mid \mathbf{f})}\!\left[\log p(\mathbf{f}\mid \mathbf{z})\right]
- D_{\text{KL}}\!\left(q(\mathbf{z}\mid \mathbf{f}) \,\|\, p(\mathbf{z})\right) \\
+ \mathbb{E}_{q(\tilde{\mathbf{z}}\mid \tilde{\mathbf{f}})}\!\left[\log p(\tilde{\mathbf{f}}\mid \tilde{\mathbf{z}})\right]
- D_{\text{KL}}\!\left(q(\tilde{\mathbf{z}}\mid \tilde{\mathbf{f}}) \,\|\, p(\tilde{\mathbf{z}})\right) \\
+ \lambda_1 \,\mathcal{L}_{\text{gan}}^{(G)} 
+ \lambda_2 \,\big\| \mathbf{1} - \boldsymbol{\mu}_n + \boldsymbol{\mu}_a \big\|_2 .
\end{multline}
Here $\mathcal{L}_{\text{gan}}^{(G)}$ is the non-saturating generator loss in feature space; $\boldsymbol{\mu}_n,\boldsymbol{\mu}_a$ are means of normal/anomaly encoders features in the current domain. As in \cite{wiewel2019continual}, terms containing $\tilde{\mathbf{f}}$ utilize features from past domains for replay and are omitted when $t{=}1$ (no replay available). %

Additionally, to enhance the separability of each class, we add a distance loss for each private latent variable.

\subsection{Multi-Discriminator}
As mentioned in the introduction, the incompleteness of anomaly detection in the WVAD framework becomes more severe under the CL setup. To address this, we employ multiple discriminators paired with DG to identify anomalies. Inspired by Zhang et al. (2023), who proposed a strategy to separate multiple head outputs, we introduce a loss that separates intermediate features among discriminators $\mathcal{D}$, $\mathcal{D}_n$, and $\mathcal{D}_a$, bringing diversity to the ensemble of models.
Specifically, for any discriminator, we optimize the following loss function.

\begin{equation}
\mathcal{L}_{\text{diver}} \;=\; \sum_{\substack{j=1 \\ j \neq k}}^{3} \big\| \mathbf{1} -  \mathbf{h}_k - \mathbf{h}_j \big\|_2 ,
\end{equation}
where $\mathbf{h}_k$ is the intermediate feature vector (just before the final layer) of the $k$-th discriminator.

In addition, inspired by the multi-task GAN~\cite{bai2018sod}, 
we install another novel connection that run from the GAN head in discriminators ($\mathcal{D}$, $\mathcal{D}_n$, and $\mathcal{D}_a$) into both the normal generator $\mathcal{G}_n$ and the anomaly generator $\mathcal{G}_a$ as we expect this allows the discriminators to handle normal and anomaly data separately and thus modules can perform more flexibly both in discriminator and generators. 
And then, this multitask strategy enables the efficient application of GAN discriminators from the DG to score ensembling, without the need to prepare multiple independent classifiers.

Specifically, we add not only a fully-connected regression layer (predicting anomaly scores) but also a fully-connected classification layer (determining true or false) behind the third layer of the discriminator network. That is, the objective of the true-false discriminator $\mathcal{D}$ is as follows:

\begin{equation} 
\mathcal{L}_{\text{gan}}^{(D)} \;=\; \mathbb{E}_{\mathbf{f} \sim p_{\text{data}}(\mathbf{f})}[\log \mathcal{D}({\mathbf{f}})] + \mathbb{E}_{\tilde{\mathbf{f}} \sim \mathcal{G}}[\log(1 - \mathcal{D}(\tilde{\mathbf{f}}))].
\end{equation}

Finally, the overall loss of the discriminator is defined as:

\begin{equation}
\mathcal{L}_\text{dis} \;=\; \mathcal{L}_{\text{MIL}} + \lambda_3 \,\mathcal{L}_{\text{gan}}^{(D)} + \lambda_4 \,\mathcal{L}_{\text{diver}}
\end{equation}
where $\lambda_3$ and $\lambda_4$ are hyperparameters for the adversarial and diver loss, respectively.

\subsection{Ensemble}
Finally, we describe how our proposed CADE works using Multi-Discriminator during inference. 
As shown in \cite{doan2023continual}, the fusion of the outputs of multiple models improves the quality of uncertain predictions. 
Also, while existing GR-based studies do not use the multiple discriminators during inference, this strategy allows for efficient utilization of GAN model discriminators used in GR and enables comprehensive anomaly detection through the use of multiple discriminators.
In the ensemble, we take the average of the scores for each model, following~\cite{han2021gan}. 
The anomaly score for the input $\mathbf{v}$, split into segments, is defined using discriminators $\mathcal{D}$,$\mathcal{D}_n$,$\mathcal{D}_a$ as follows.
\begin{equation}
a(\mathbf{v}) \;=\; \frac{1}{N}\Big\{\mathcal{D}(\mathcal{I}(\mathbf{v}))+\mathcal{D}_n(\mathcal{I}(\mathbf{v})) + \mathcal{D}_a(\mathcal{I}(\mathbf{v}))\Big\}
\end{equation}
where $N$ is the total number of models used in the ensemble.

In summary, CADE consists of three components --- Dual-Generator (DG), Multi-Discriminator (MD), and Ensemble --- that work in complementary ways to let the WVAD framework effectively overcome forgetting.

%% file: sec/4_exp.tex
\section{Experiments}
\label{sec:exp}

\newcommand{\stdpm}[2]{#1{\scriptsize$\pm$#2}}
\newcommand{\best}[2]{\textbf{#1}{\scriptsize$\pm$#2}}

\noindent
\textbf{Datasets}:
Proposed CADE is evaluated on three multi-scene datasets, ShanghaiTech (SHT) dataset~\cite{liu2018future} , Charlotte Anomaly Dataset (CHAD) dataset~\cite{danesh2023chad} and UCF-Crime dataset~\cite{sultani2018real} which are most suitable for the realistic problem setting of VAD applications with changing scenes.

ShanghaiTech is a medium-sized dataset containing 437 videos taken at 13 different scenes, 307 of which are normal and 130 of which contain anomalies. 
In this paper, we use another modified dataset especially designed for weakly-supervised evaluation, following \cite{zhang2019temporal}. 
For evaluation with this dataset in DIL manner, we define each shooting scene as domain (i.e., 13 domains are involved) and divide data of each domain into 10 parts.
Meanwhile, the CHAD contains 412 videos filmed in four different scenes. The domains correspond directly to these scenes, and we evaluate CADE across these four scenarios.
UCF\mbox{-}Crime contains 1{,}610 training videos with video\mbox{-}level labels and 290 test videos with frame\mbox{-}level annotations~\cite{sultani2018real}. For DIL evaluation, we cluster video\mbox{-}level CLIP embeddings of the training videos (no labels or prompts) into 10 scene\mbox{-}like domains and introduce them sequentially, while keeping the official train/test split fixed. In practice, per\mbox{-}camera streams at a site naturally form such homogeneous scene clusters, so this setup reflects real deployment. Per\mbox{-}domain examples and implementation details are provided in the Supplementary.

\noindent
\textbf{Metrics}:
Following previous studies~\cite{feng2021mist,sultani2018real,tian2021weakly}, we compute the area under the curve (AUC) of the frame-level receiver operating characteristic (ROC) as the main metric. To evaluate the degree of forgetting, we also compute and evaluate the final accuracy for all test data, i.e., AUROC, after the model has been trained for all domains.

\noindent
\textbf{Baselines}:
As mentioned in the introduction, CADE can be applied to existing WVAD methods. We choose four recent methods, \textbf{MIST}~\cite{feng2021mist}, \textbf{Sultani et al.}~\cite{sultani2018real}, \textbf{RTFM}~\cite{tian2021weakly}, and \textbf{UR-DMU}~\cite{URDMU_zh}; we apply CADE to these methods to investigate whether enabling CADE leads to performance gain.
Note that the configuration of the discriminator $\mathcal{D}$,$\mathcal{D}_n$,$\mathcal{D}_a$ is based on each method, but a GAN header is added to the discriminator for the GAN framework of the generator and discriminator as described in the section of proposal.
We consider, for each WVAD backbone, the following training regimes: (1) \textbf{FT} — continual fine\mbox{-}tuning in the DIL setup; (2) \textbf{EWC}~\cite{Kirkpatrick2017OvercomingCF} — elastic weight consolidation; (3) \textbf{SI}~\cite{Zenke2017SI} — synaptic intelligence; (4) \textbf{iCaRL}~\cite{rebuffi2017icarl} — exemplar replay; (5) \textbf{w/\,CADE} — our method; and (6) \textbf{MTL} — joint training on all domains. EWC/SI/iCaRL are implemented following recommended settings, with training iterations, batch sizes matched across methods for fairness. Note that comparisons to continual learning baselines (EWC/SI/iCaRL) are provided only on UCF\mbox{-}Crime. As is common in CL, \textbf{MTL} does not follow DIL and serves as an oracle upper bound for reference. 
Note that, for MIST~\cite{feng2021mist}, although it employs a two-stage training process, only the first stage of the MIL-based training process is used for verification. %
Note that comparisons to continual learning baselines (EWC/SI/iCaRL) are provided only on UCF\mbox{-}Crime.

Although UR-DMU is not an MIL-based method, we introduced CADE with regarding its classifier that includes multiple components as a discriminator.

\noindent
\textbf{Training Details}:
We follow the feature extraction methods of the respective baseline approaches. Specifically, Sultani uses C3D~\cite{tran2015learning}, MIST and UR-DMU use I3D~\cite{carreira2017quo}, and RTFM uses 10-crop augmented I3D. The segment length is set to 16. Training is conducted for 10 epochs per domain. The DG has an encoder/decoder consisting of a 3-layer MLP and is optimized by Adam with a learning rate of 0.001.
The experiments are conducted three times with different seeds, and the average of AUC and standard deviation are reported.

\begin{table*}[ht]
\centering
\caption{The final AUC and variance for SHT and CHAD. Best methods in each dataset are emphasized in \textbf{bold}.}
\resizebox{\textwidth}{!}{%
\begin{tabular}{@{}lcccccc@{}}
\toprule
 & \multicolumn{3}{c}{SHT} & \multicolumn{3}{c}{CHAD} \\ \cmidrule(lr){2-4} \cmidrule(lr){5-7}
 & FT & w/CADE & MTL & FT & w/CADE & MTL \\ \midrule
Sultani et al. & 0.5987 \newline {\scriptsize$\pm$ 0.0571} & \textbf{0.6811} \newline {\scriptsize$\pm$ 0.0052} & 0.7850 \newline {\scriptsize$\pm$ 0.0201} & 0.5872 \newline {\scriptsize$\pm$ 0.0156} & \textbf{0.6694} \newline {\scriptsize$\pm$ 0.0424} & 0.7160 \newline {\scriptsize$\pm$ 0.0145}\\
MIST & 0.5640 \newline {\scriptsize$\pm$ 0.1118} & \textbf{0.8490} \newline {\scriptsize$\pm$ 0.0057} & 0.8515 \newline {\scriptsize$\pm$ 0.0161} & 0.6768 \newline {\scriptsize$\pm$ 0.0202} & \textbf{0.7674} \newline {\scriptsize$\pm$ 0.0058} & 0.7732 \newline {\scriptsize$\pm$ 0.0118} \\
RTFM & 0.4044 \newline {\scriptsize$\pm$ 0.0482} & \textbf{0.7474}
\newline {\scriptsize$\pm$ 0.0246} & 0.9001
 \newline {\scriptsize$\pm$ 0.0077} & 0.5684 \newline {\scriptsize$\pm$ 0.0489} & \textbf{0.7318} \newline {\scriptsize$\pm$ 0.0019}  & 0.8101 \newline {\scriptsize$\pm$ 0.0046} \\
UR-DMU & 0.7984  \newline {\scriptsize$\pm$ 0.0116} & \textbf{0.8483} \newline {\scriptsize$\pm$ 0.0141} & 0.8567 \newline {\scriptsize$\pm$ 0.0331} & 0.7066 \newline {\scriptsize$\pm$ 0.0539} & \textbf{0.7684} \newline {\scriptsize$\pm$ 0.0233} & 0.8000 \newline {\scriptsize$\pm$ 0.0024} \\ \bottomrule
\end{tabular}%
}
\label{tab:accuracy}
\end{table*}

\begin{table*}[ht]
  \centering
  \caption{The final AUC and variance for UCF-Crime. Best methods are emphasized in \textbf{bold}.}
  \label{tab:experiment-results}
  \renewcommand{\arraystretch}{0.8}
  \setlength{\tabcolsep}{5pt}
  \begin{adjustbox}{width=\textwidth}
    \begin{tabular}{@{}lcccccc@{}}
      \toprule
      \textbf{} & FT & EWC & SI & iCaRL & w/\,CADE & (MTL) \\
      \midrule
      Sultani et al. & \stdpm{0.701}{0.006} & \stdpm{0.735}{0.009} & \stdpm{0.729}{0.009} & \stdpm{0.733}{0.008} & \best{0.743}{0.008} & \stdpm{0.769}{0.002} \\
      RTFM           & \stdpm{0.445}{0.051} & \stdpm{0.691}{0.068} & \stdpm{0.709}{0.024} &         \stdpm{0.690}{0.037}    & \best{0.800}{0.003} & \stdpm{0.865}{0.020} \\
      MIST           & \stdpm{0.738}{0.003} & \stdpm{0.739}{0.004} & \stdpm{0.742}{0.004} & \stdpm{0.737}{0.004} & \best{0.743}{0.004} & \stdpm{0.765}{0.002} \\
      UR\mbox{-}DMU  & \stdpm{0.744}{0.005} & \stdpm{0.758}{0.008} & \stdpm{0.745}{0.006} & \stdpm{0.764}{0.016} & \best{0.793}{0.006} & \stdpm{0.798}{0.003} \\
      \bottomrule
    \end{tabular}
  \end{adjustbox}
\end{table*}

\begin{table}[ht]
\centering
\caption{Ablation results on SHT and CHAD datasets.}
\resizebox{\columnwidth}{!}{%
\begin{tabular}{c c c c c c c}
\toprule
 & (1) & (2)  & (3) & (4) & SHT & CHAD  \\ \midrule
 & \checkmark &  &  &  & 0.6926 {\scriptsize$\pm$ 0.1857} & 0.5364 {\scriptsize$\pm$ 0.1041} \\
 & & \checkmark &  &  & 0.7980 {\scriptsize$\pm$ 0.0188} & 0.6869 {\scriptsize$\pm$ 0.0295} \\
 &  & \checkmark & \checkmark &  & 0.8039 {\scriptsize$\pm$ 0.0169} & 0.7126 {\scriptsize$\pm$ 0.0187} \\
 &  & \checkmark & \checkmark & \checkmark & \textbf{0.8490} {\scriptsize$\pm$ 0.0057} & \textbf{0.7674} {\scriptsize$\pm$ 0.0058} \\ \bottomrule
\end{tabular}
}
\label{tab2}
\end{table}

\begin{table}[ht]
\centering
\caption{Comparison of replay size on SHT and CHAD.}
\resizebox{\columnwidth}{!}{%
\begin{tabular}{ccccccc}
\toprule
{Replay Size} & 1/3 & 2/3 & 1 & 4/3 & 5/3 & 6/3 \\ \midrule
\multirow{2}{*}{SHT} & 0.7854 & 0.8336 & \textbf{0.8490} & 0.8131 & 0.8176 & 0.8094 \\
 & \scriptsize$\pm$ 0.0050 & \scriptsize$\pm$ 0.0099 & \scriptsize$\pm$ 0.0057 & \scriptsize$\pm$ 0.0042 & \scriptsize$\pm$ 0.0245 & \scriptsize$\pm$ 0.0209 \\ \midrule
\multirow{2}{*}{CHAD} & 0.7642 & 0.7619 & 0.7663 & 0.7585 & \textbf{0.7674} & 0.7611 \\
 & \scriptsize$\pm$ 0.0026 & \scriptsize$\pm$ 0.0098 & \scriptsize$\pm$ 0.0035 & \scriptsize$\pm$ 0.0161 & \scriptsize$\pm$ 0.0058 & \scriptsize$\pm$ 0.0092 \\ \bottomrule
\end{tabular}
}
\label{tab3}
\end{table}

\begin{table}[ht]
\centering
\caption{Comparison of scene on SHT, CHAD and UCF-Crime.}
\resizebox{\columnwidth}{!}{%
\begin{tabular}{c c c c c c}
\toprule
Scene & 2 & 4 & 6 & 8 & 10 \\ 
\midrule
\multirow{2}{*}{SHT} 
   & 0.6205 & 0.7358 & 0.8215 & 0.8208 & \textbf{0.8490} \\
   & {\scriptsize$\pm$0.0052} & {\scriptsize$\pm$0.0252} & {\scriptsize$\pm$0.0214} & {\scriptsize$\pm$0.0115} & {\scriptsize$\pm$0.0057} \\
\midrule
Scene & 1 & 2 & 3 & 4 &  \\
\midrule
\multirow{2}{*}{CHAD} 
   & 0.7494 & 0.7373 & 0.7505 & \textbf{0.7674} & \\
   & {\scriptsize$\pm$0.0181} & {\scriptsize$\pm$0.0105} & {\scriptsize$\pm$0.0161} & {\scriptsize$\pm$0.0058} & \\
\midrule
Scene & 2 & 4 & 6 & 8 & 10 \\ 
\midrule
\multirow{2}{*}{UCF-Crime} 
   & 0.5393 & 0.7347 & 0.7547 & 0.7927 & \textbf{0.7934} \\
   & {\scriptsize$\pm$0.0518} & {\scriptsize$\pm$0.0169} & {\scriptsize$\pm$0.0217} & {\scriptsize$\pm$0.0144} & {\scriptsize$\pm$0.0090} \\
\bottomrule
\end{tabular}
}

\label{tab4}
\end{table}

\subsection{Results on ShanghaiTech (SHT)}

\cref{tab:accuracy} shows the final AUC, indicating that CADE outperforms FT results for all existing methods.
On the other hand, CADE underperforms MTL, which is as expected because MTL does not follow DIL setup and MTL fully exploits all domain data together. However, MIST with CADE is only 0.0025 of AUC worse than MTL, which demonstrates the effectiveness of CADE.
In the FT, the model is strongly affected by destructive forgetting and even performs as same as randomly in Sultani et al., MIST and RTFM case (i.e., AUC score is 0.5).
On the other hand, UR-DMU is less affected by forgetting (e.g., the AUC score is 0.7984). This is likely due to the Dual-Memory Unit that retains several prototypes internally, helping to mitigate the forgetting of past scenes.

\begin{figure}[!htbp]
\centering
\begin{minipage}[b]{0.48\linewidth}
  \centering
  \includegraphics[width=\textwidth]{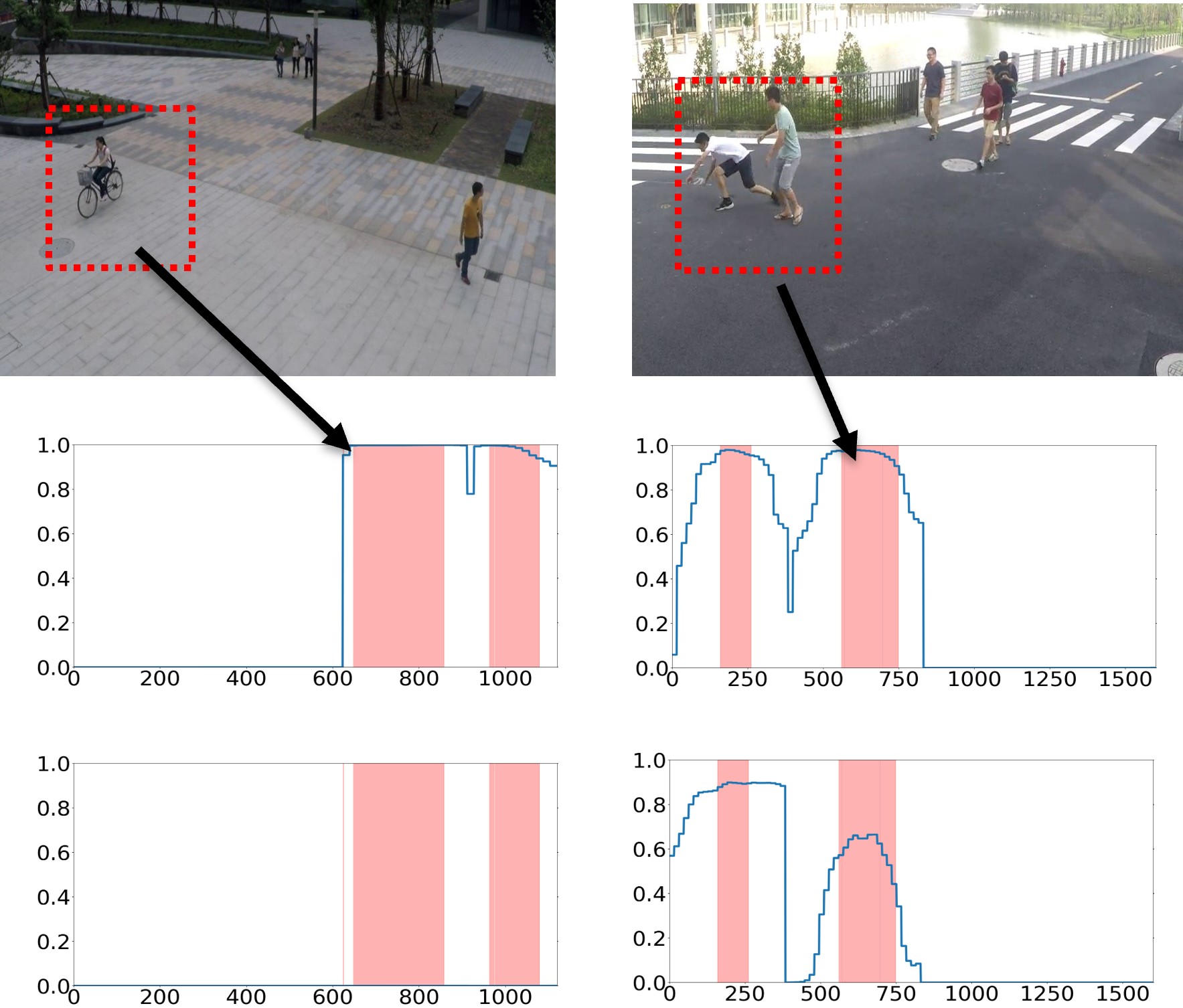}
  \subcaption{SHT results}\label{fig:sht}
\end{minipage}
\hfill
\begin{minipage}[b]{0.48\linewidth}
  \centering
  \includegraphics[width=\textwidth]{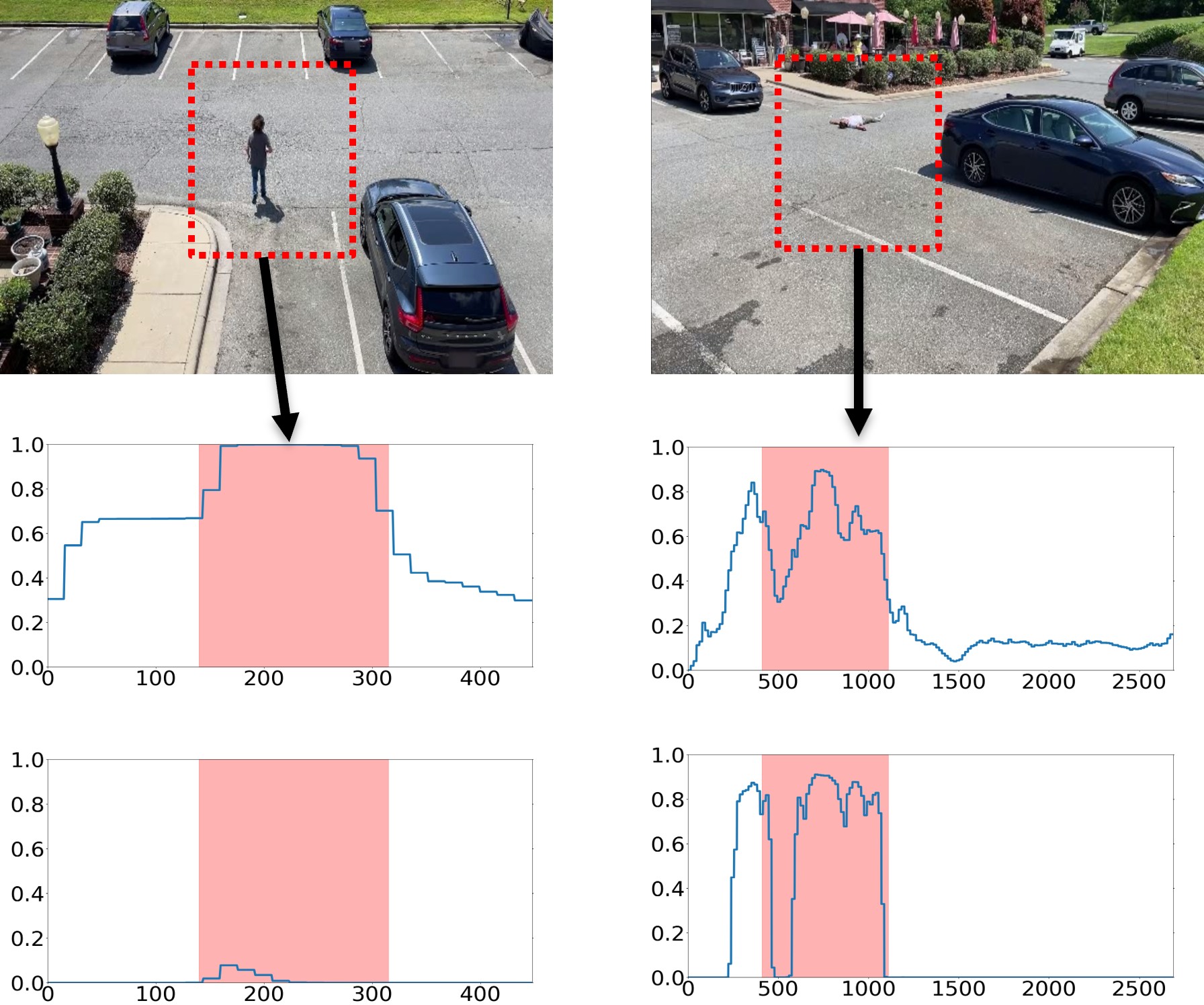}
  \subcaption{CHAD results}\label{fig:chad}
\end{minipage}
\caption{Qualitative evaluation on SHT and CHAD datasets. Red indicates anomaly annotations, blue shows the score. Left columns represent past scenes, right columns recent scenes. The top row shows CADE results, the bottom row shows fine-tuned UR-DMU results.}
\label{fig3}
\end{figure}

\cref{fig3} (a) shows the qualitative evaluation. 
In the fine-tuning case, the sensitivity to anomalies in past scenes (left column) significantly decreases, indicating that UR-DMU still suffers from forgetting.
These results clearly support that our CADE contributes to the forgetting prevention.

\subsection{Results on Charlotte Anomaly Dataset (CHAD)}
Despite consisting of relatively short 4 scenarios, CHAD still shows that existing methods suffer from forgetting, as shown in \cref{tab:accuracy} and \cref{fig3}(b). However, CADE also outperforms FT results for all existing methods on CHAD. Similarly to SHT, MIST with CADE is only 0.0058 AUC worse than MTL. This indicates that CADE is effective across multiple datasets and methods.

\subsection{Results on UCF-Crime Dataset}
Table~2 reports the evaluation on UCF\mbox{-}Crime. Across all backbones, CADE consistently surpasses FT and achieves higher AUROC than representative continual learning baselines (EWC/SI/iCaRL). Concretely, with RTFM CADE attains $0.800{\pm}0.003$, with UR\mbox{-}DMU $0.793{\pm}0.006$, and with MIST $0.743{\pm}0.004$, outperforming FT (e.g., RTFM $0.445{\pm}0.051$) as well as regularization/rehearsal\mbox{-}based methods. This trend indicates that, under the weak labels and rarity of anomalies characteristic of WVAD, CADE’s design—namely generative replay via the dual generator and the multi\mbox{-}discriminator ensemble—effectively addresses these challenges, whereas existing CL methods remain limited in this setting.
Furthermore, the domain\mbox{-}sensitivity analysis in Table~5 shows that AUROC increases monotonically as scenes progress (e.g., from $0.5393$ at Scene~2 to $0.7934$ at Scene~10), evidencing effective suppression of forgetting of past\mbox{-}domain knowledge. This mirrors the SHT results and supports that CADE is particularly effective when anomaly patterns differ across scenes.

\subsection{Ablation Studies}
Since CADE incorporates several components to achieve superior performance, we conducted additional evaluations to assess their effectiveness.
Specifically, we observe the change in performance with different combinations of the following four components. MIST is used as the baseline in all settings.
(1) VAE is attached to the baseline (i.e., MIST).
(2) Dual-Generator is attached.
(3) Multi-Discriminator is attached.
(4) Distance loss component is considered.
(1) is the VAE-based CL method most closely aligned with our motivation, using two VAEs to handle video-level labels. 
Table 3 represents the combinations and results. Note that the bottom combination that includes (2), (3), and (4) corresponds to the proposed CADE itself.
As shown in \cref{tab2}, CADE that combines all components achieves up to 0.1564 improvement in AUC on the SHT dataset compared to the baseline.
The DG alone improves AUC by 0.1054 on the SHT dataset compared to the baseline. Particularly on CHAD, existing VAE baseline methods fail to effectively prevent forgetting, likely due to the longer video clips and the rarity and imbalance of anomalies compared to SHT. The DG effectively addresses these characteristics. The combination of MD and distance loss is even more effective, with the distance loss contributing an additional 0.0451 improvement in AUC on SHT.
This demonstrates that CADE compensates for the uncertainty in the discriminators' predictions due to forgetting, and captures the incompleteness of anomalies through the diverse outputs of multiple discriminators.
Additionally, as shown in \cref{tab3}, we evaluated the performance of the CADE architecture with different replay sizes and scene transitions. For the SHT dataset, a generation ratio of 50 percent relative to the raw data is desirable. Meanwhile, for the CHAD dataset, the replay ratio does not significantly affect the performance.
Furthermore, as shown in \cref{tab4}, the AUC increases almost monotonically with scene changes, indicating that knowledge transfer in CADE is functioning effectively. The high performance in the initial scenes of the CHAD dataset, unlike SHT, is likely due to the commonality of anomaly scenarios across scenes.

%% file: sec/5_conc.tex
\section{Conclusion}
\label{sec:conc}

We propose a novel WVAD method that can handle continual learning scenarios, CADE, which is the first paper to incorporate a CL problem set into WVAD. The generative replay significantly mitigates the forgetting of classifiers. In addition, we propose to ensemble discriminators and generators during inference to compensate for the loss of prediction quality due to classifier forgetting. Extensive experiments demonstrate that CADE significantly outperforms existing WVAD methods on common WVAD benchmarks, such as the ShanghaiTech, UCF-Crime, and Charlotte Anomaly Dataset.

\noindent
\textbf{Acknowledgment.}
This work was partly supported by the Ministry of Internal Affairs and Communications of Japan under project JPMI00316.

%% file: sec/6_supp.tex
\makeatletter\setlength{\@fptop}{0pt}\setlength{\@dblfptop}{0pt}\makeatother
\renewcommand{\floatpagefraction}{0.5}
\subsection*{ShanghaiTech Dataset Setup}
\begin{figure}[h]
\centering
\begin{minipage}[h]{\linewidth}
  \centering
  \includegraphics[width=\textwidth]{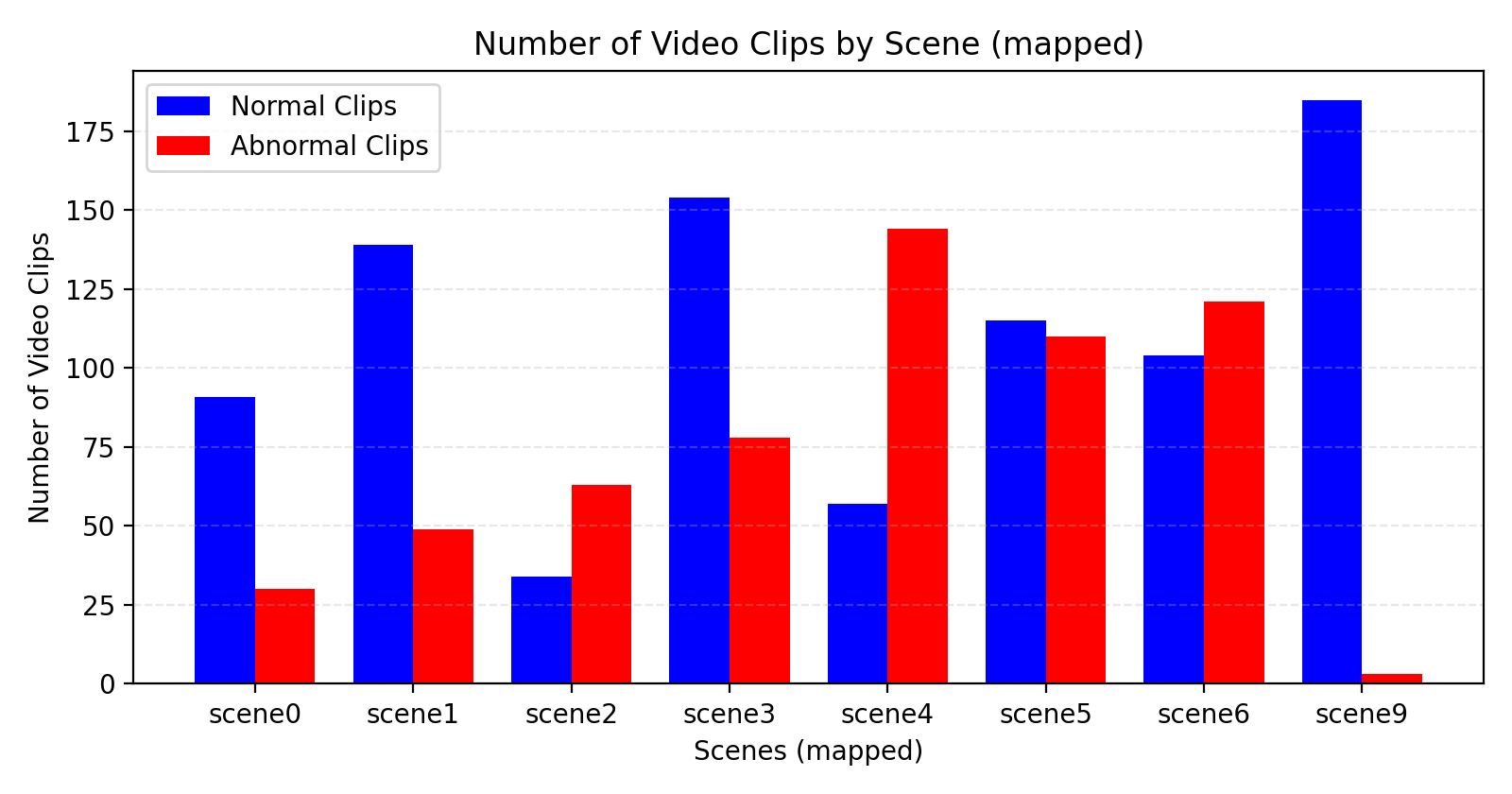}
\end{minipage}
\caption{Number of Video Clips per Scene in the Reorganized ShanghaiTech Dataset.
}
\label{supp:fig:sht_clips}
\end{figure}

\begin{figure}[h]
\centering
\begin{minipage}[h]{\linewidth}
  \centering
  \includegraphics[width=\textwidth]{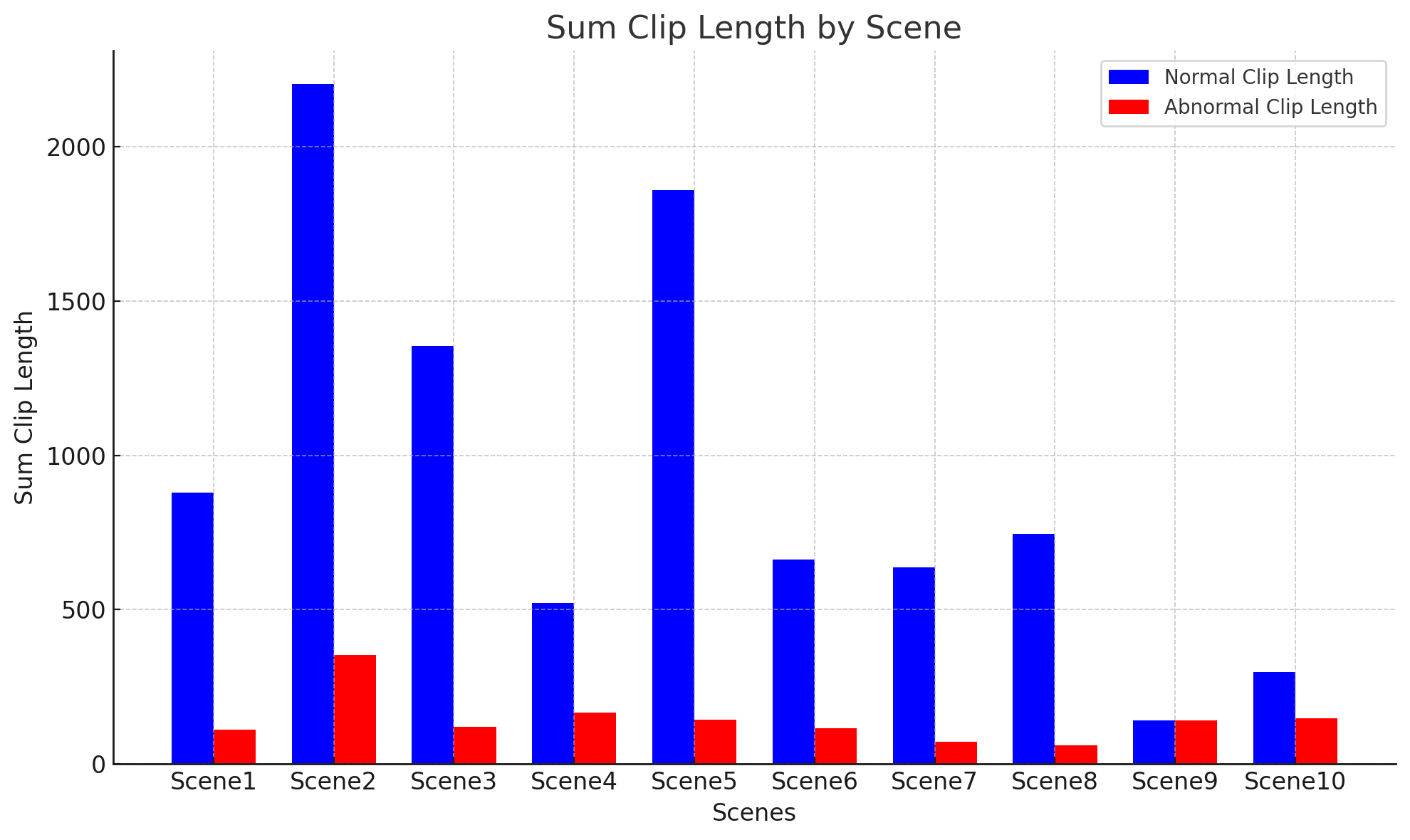}
\end{minipage}
\caption{Total Length of Video Clips per Scene in the Reorganized ShanghaiTech Dataset.}
\label{supp:fig:sht_len}
\end{figure}

\begin{figure}[p]
\centering
\begin{minipage}[h]{\linewidth}
  \centering
  \includegraphics[width=\textwidth]{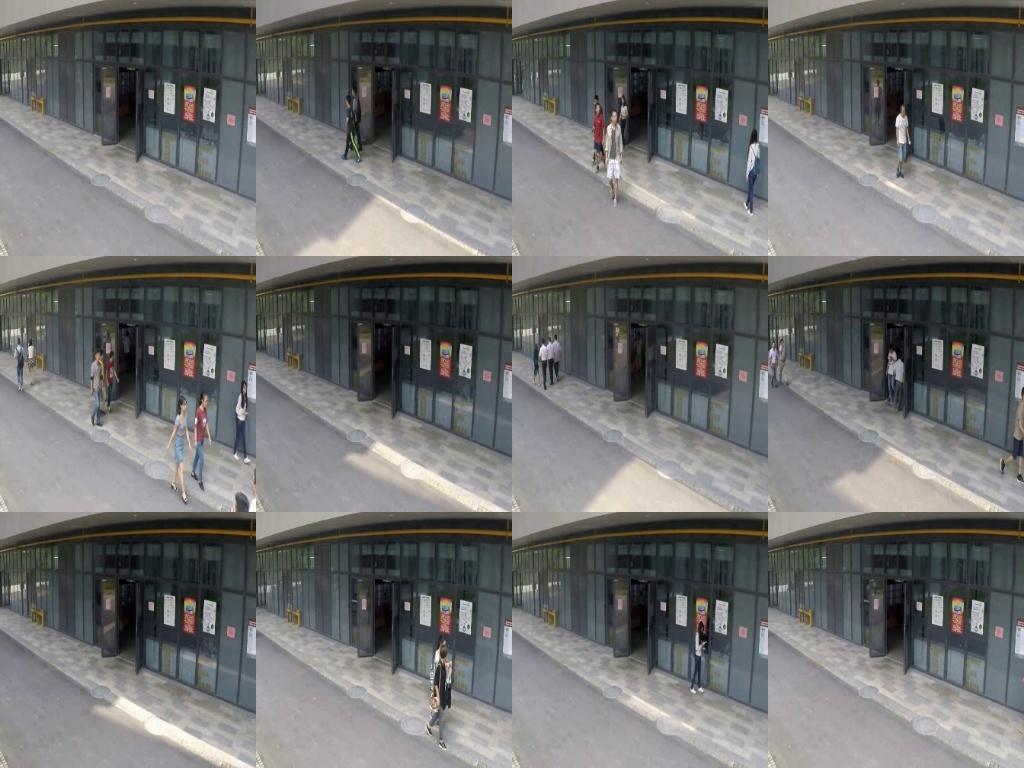}
\end{minipage}
\caption{Example frames sampled from the divided domains of SHT (Scene 8).}
\label{supp:fig:sht_ex8}
\end{figure}

\begin{figure}[p]
\centering
\begin{minipage}[h]{\linewidth}
  \centering
  \includegraphics[width=\textwidth]{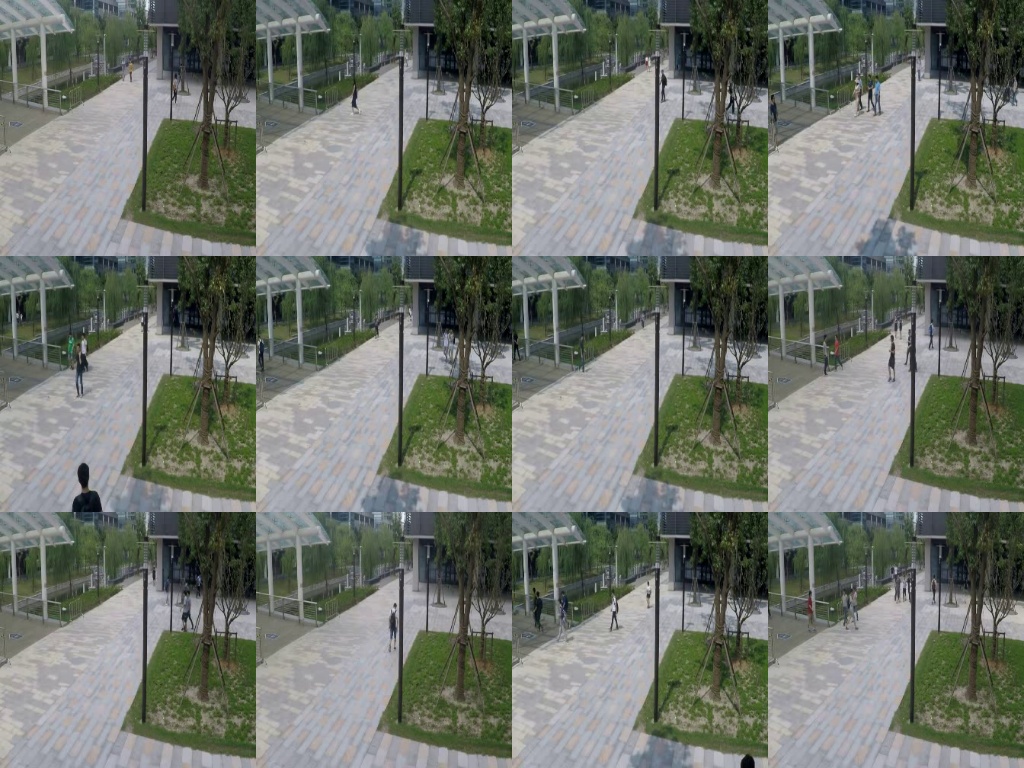}
\end{minipage}
\caption{TExample frames sampled from the divided domains of SHT (Scene 3).}
\label{supp:fig:sht_ex3}
\end{figure}

To validate our CADE framework, we explain the scenario splitting method used for the ShanghaiTech dataset~\cite{liu2018future}. Originally, ShanghaiTech is a multi-scene dataset consisting of 13 scenes, but due to the imbalance in the amount of data across scenes, we re-divided it into 10 scenes. Specifically, the reorganization is shown in \cref{supp:tab:mapping}. Additionally, the distribution of the amount of training data in each scene after reorganization is presented in \cref{supp:fig:sht_clips} and \cref{supp:fig:sht_len}. The divided scenes are mutually independent and non-overlapping. Examples of the divided scenes are shown in \cref{supp:fig:sht_ex8} and \cref{supp:fig:sht_ex3}. Training and test splits follow the official protocol.

\begin{table}[h]
\centering
\caption{Scene Mapping}
\resizebox{\columnwidth}{!}{%
\begin{tabular}{c c c c c c c c c c c c c c}
\toprule
\textbf{Original} & \textbf{01} & \textbf{02} & \textbf{03} & \textbf{04} & \textbf{05} & \textbf{06} & \textbf{07} & \textbf{08} & \textbf{09} & \textbf{10} & \textbf{11} & \textbf{12} & \textbf{13} \\
\midrule
\textbf{Reordered} & 2 & 7 & 10 & 4 & 5 & 8 & 9 & 3 & 7 & 6 & 6 & 1 & 10 \\
\bottomrule
\end{tabular}
}
\label{supp:tab:mapping}
\end{table}

\subsection*{UCF-Crime Dataset Setup}
To construct domain-incremental learning (DIL) settings on UCF-Crime~\cite{sultani2018real}, we divided the dataset into ten domains in an unsupervised manner without using labels or textual prompts. For each video, multiple frames were uniformly sampled after discarding the initial portion, and frames with extremely low quality (e.g., low contrast or nearly blank) were removed. Each frame was embedded using CLIP, and video-level representations were obtained by aggregating frame embeddings in a robust manner. These video embeddings were then clustered with $k$-means ($k=10$), and the resulting clusters were treated as distinct domains.

\begin{figure}[p]
\centering
\begin{minipage}[h]{\linewidth}
  \centering
  \includegraphics[width=\textwidth]{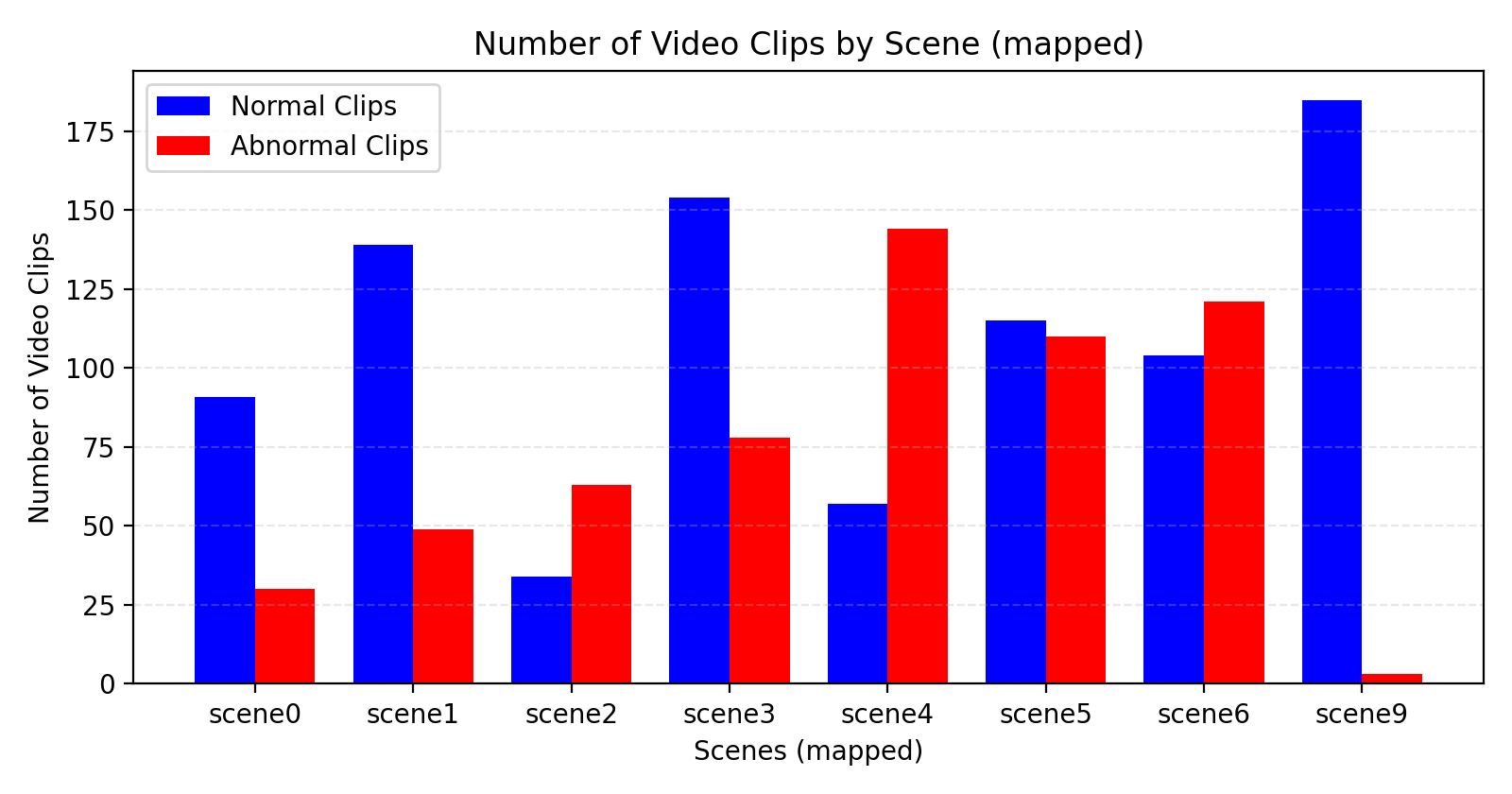}
\end{minipage}
\caption{Number of Video Clips per Scene in the Reorganized UCF-Crime Dataset.
}
\label{supp:fig:ucf_clips}
\end{figure}

\begin{figure}[p] %
\centering
\includegraphics[width=\columnwidth]{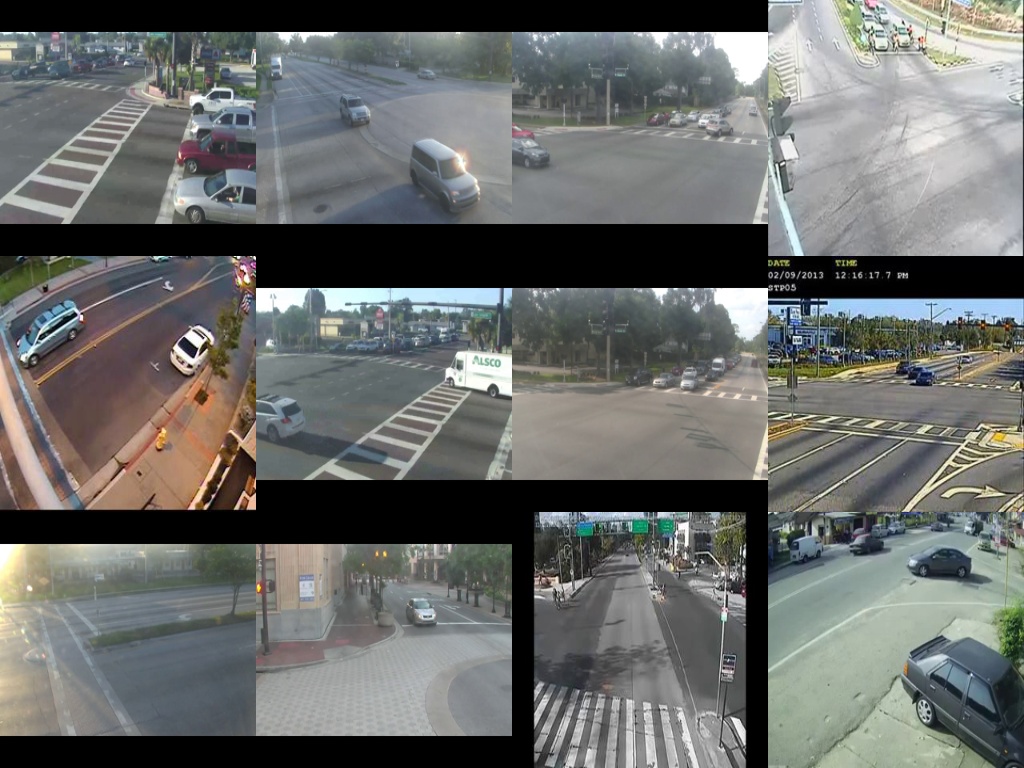} %
\caption{Example frames sampled from the divided domains of UCF-Crime. Scene 3 mainly forms a cluster of outdoor road scenes.}
\label{supp:fig:ucf_ex3}
\end{figure}

\begin{figure}[p] %
\centering
\includegraphics[width=\columnwidth]{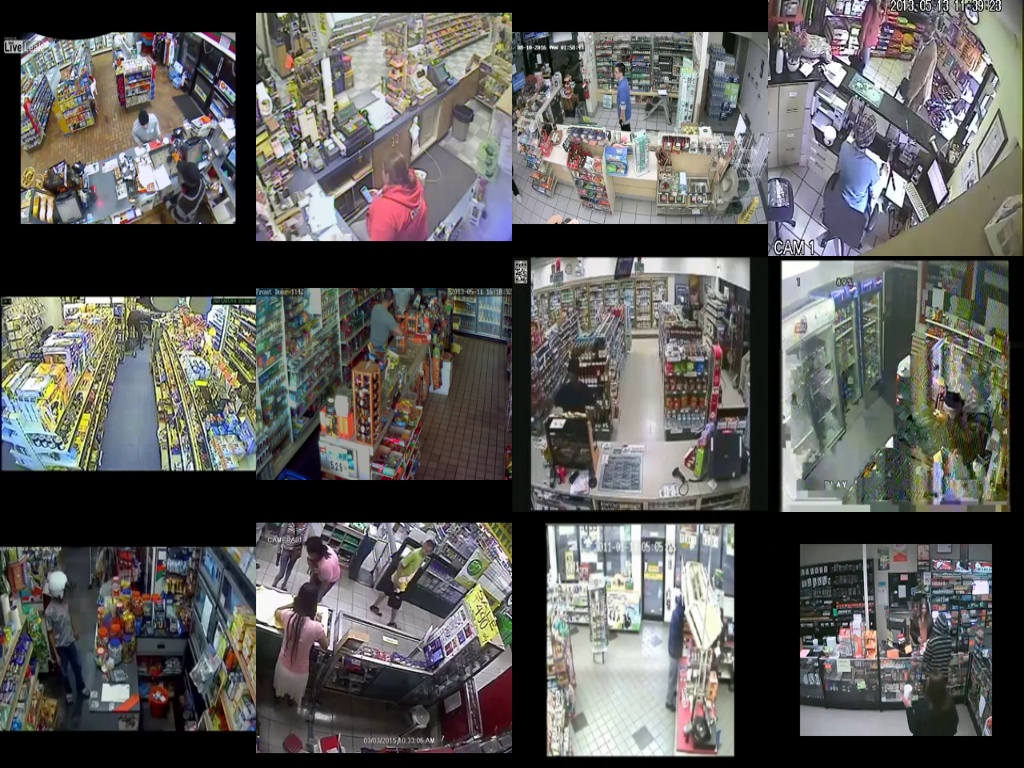} %
\caption{Example frames sampled from the divided domains of UCF-Crime. Scene 4 mainly forms a cluster of outdoor road scenes.}
\label{supp:fig:ucf_ex4}
\end{figure}

As illustrated in \cref{supp:fig:ucf_ex3}, \cref{supp:fig:ucf_ex4} , this procedure naturally yields clusters corresponding to different scene types. For example, domains emerge that predominantly capture indoor environments, outdoor public areas, or parking lots. This confirms that the unsupervised clustering effectively organizes UCF-Crime videos into meaningful scene-based domains, aligning well with the assumptions of domain-incremental evaluation.
Additionally, the distribution of the amount of data in each scene is presented in \cref{supp:fig:ucf_clips}.

\begin{table}[p]
\centering
\caption{Comparison of parameter counts and inference speed for UR-DMU with CADE.}
\label{supp:tab:efficiency}
\begin{tabular}{lcc}
\toprule
 & Parameters & Mean FPS \\
\midrule
Single Discriminator & 8,065,922  & 119.22 \\
3$\times$ Discriminator Ensemble & 24,197,766 & 117.66 \\
\bottomrule
\end{tabular}
\end{table}

\begin{figure*}[t] %
\centering
\includegraphics[width=0.88\textwidth]{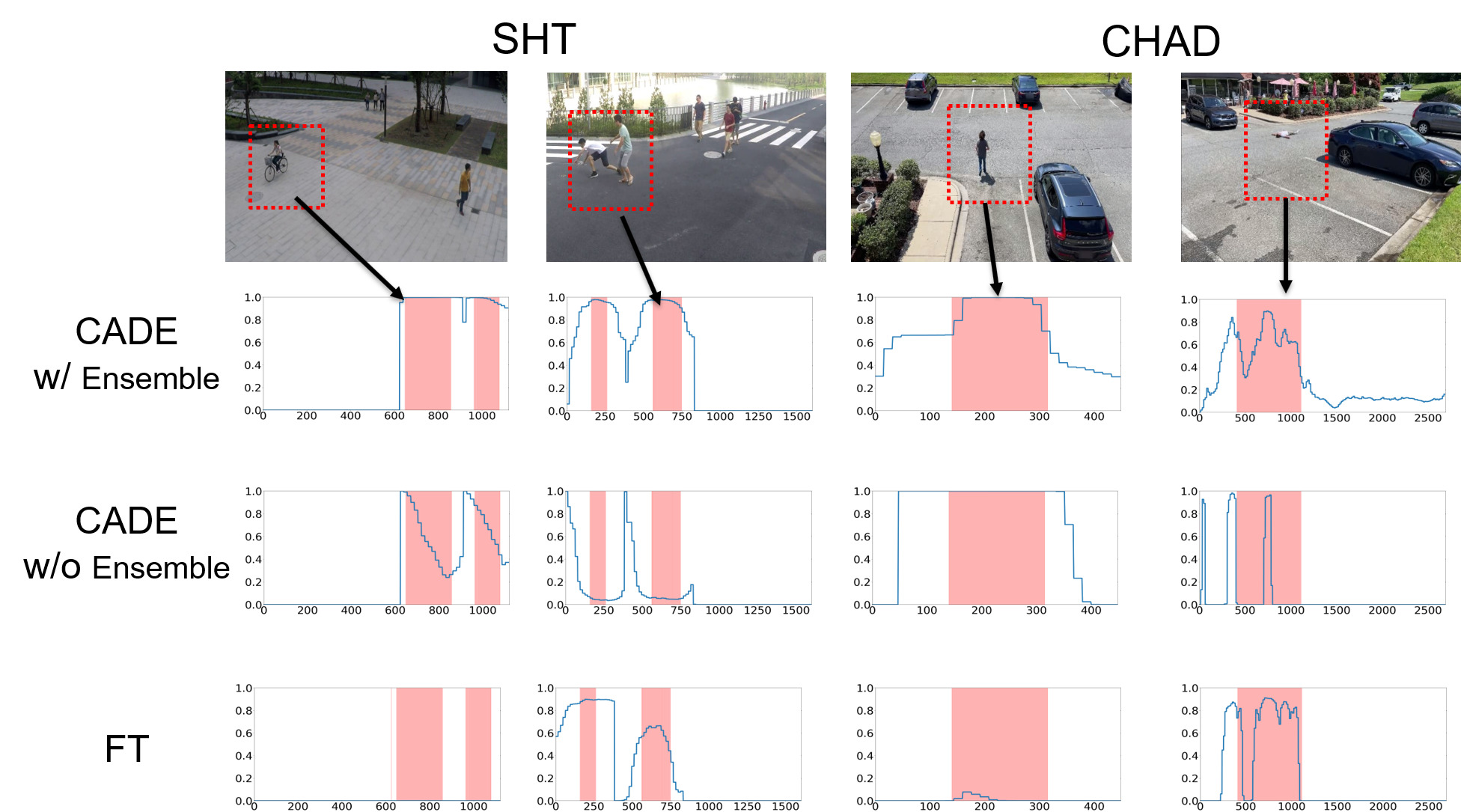}
\caption{Qualitative evaluation on UR-DMU with CADE. Red indicates anomaly annotations, blue shows the score. Each left column represents past scenes, right columns recent scenes. The top row shows CADE with ensemble results, the middle row shows CADE without ensemble results, and the bottom row shows fine-tuned UR-DMU results.}
\label{supp:fig:qual_urdmu}
\end{figure*}

\begin{figure*}[t] %
\centering
\includegraphics[width=0.88\textwidth]{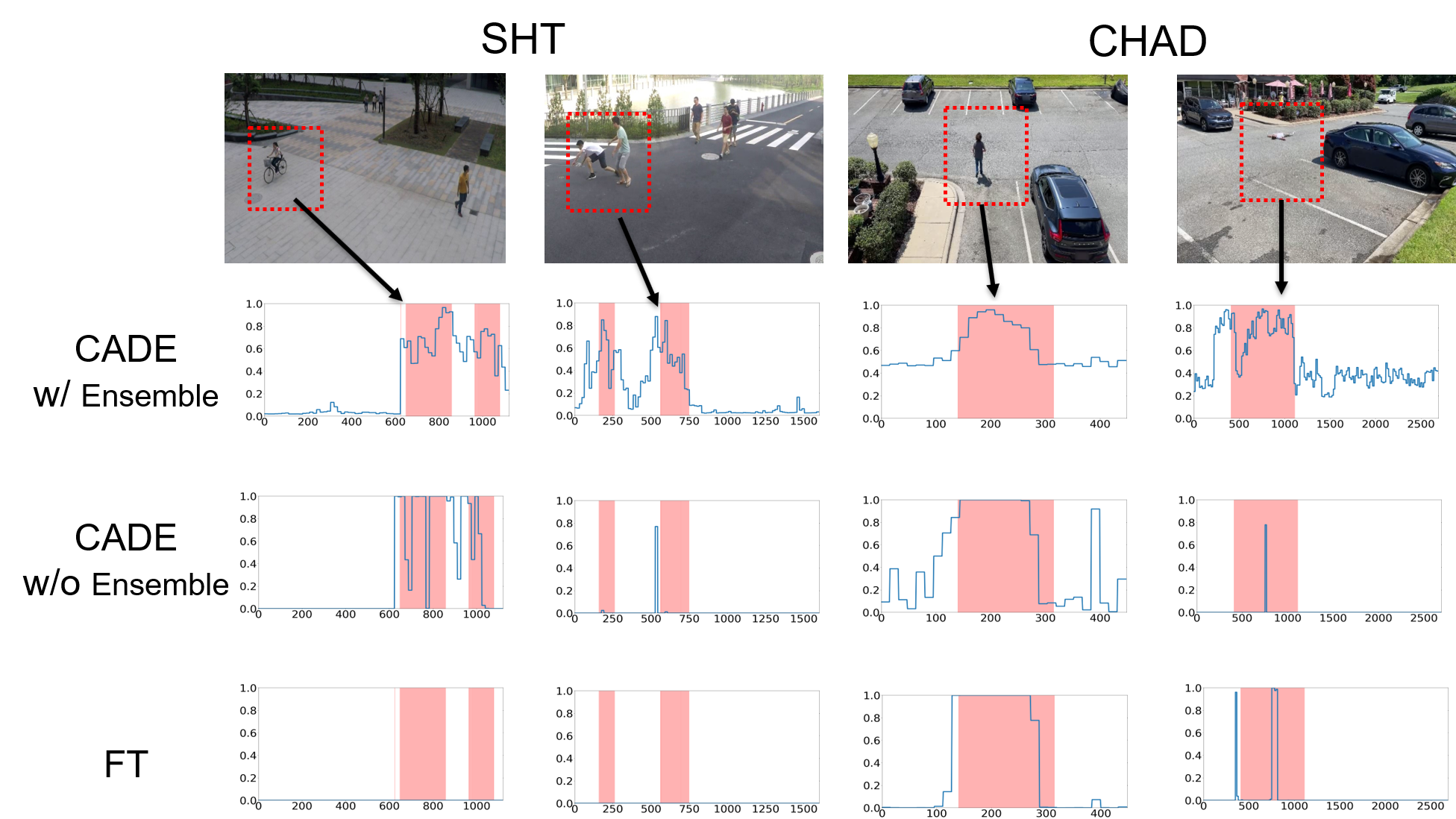}
\caption{Qualitative evaluation on MIST with CADE.}
\label{supp:fig:qual_mist}
\end{figure*}

\subsection*{Impact of Ensemble}
\cref{supp:fig:qual_urdmu} and \cref{supp:fig:qual_mist} present the qualitative evaluation results. It can be seen that our proposed CADE demonstrates high sensitivity to anomalies, regardless of the scene, whether applied to UR-DMU~\cite{URDMU_zh} in Figure~\ref{supp:fig:qual_urdmu} or MIST~\cite{feng2021mist} in Figure~\ref{supp:fig:qual_mist}. The top row of each figure shows the results with the ensemble. These results indicate a high anomaly sensitivity across both past and current scenes. The middle row shows the results without the ensemble, using only the Dual-Generator and a single discriminator. In this case, detection is incomplete, with false negatives and temporal biases evident, a phenomenon we refer to as 'incompleteness.' The bottom row, showing the fine-tuned results, reveals that forgetting in past scenes becomes increasingly severe, leading to missed detections. Thus, our CADE framework can be attached to existing methods like UR-DMU and MIST and overcomes the anomaly 'incompleteness' due to forgetting through the ensemble of Multi-Discriminator.

\subsection*{Practicality and Efficiency Evaluation}
To examine the practical feasibility of CADE, we integrated it into UR-DMU and compared the inference efficiency between a single discriminator and a multi-discriminator ensemble (3 discriminators). For the efficiency measurement, we report the average end-to-end inference time computed over 10 video clips. The results are summarized in Tab.~\ref{supp:tab:efficiency}. The parameter count increases linearly with the number of discriminators (about threefold when using three discriminators), yet the mean inference speed shows negligible difference: 119.22 FPS for a single discriminator versus 117.66 FPS for three discriminators. This is because the computational bottleneck lies in feature extraction (e.g., I3D), while the MIL-based discriminators commonly used in WVAD are lightweight. Hence, the additional cost introduced by CADE is minor in both runtime and model size.